\documentclass[11pt]{article} 
\usepackage{geometry}
\usepackage{epic,eepic}
\usepackage[parfill]{parskip} 
\usepackage{verbatim} 
\usepackage{theorem}
\usepackage{tabularx}
\usepackage{amsmath}
\usepackage{amssymb}
\usepackage{dcolumn}
\usepackage{multirow}
\usepackage{float}
\floatstyle{ruled}
\newfloat{algo}{ht}{loa}[section]
\floatname{algo}{\large{Algorithm}}

{\theorembodyfont{\rmfamily}} \theoremstyle{break}
{\theorembodyfont{\rmfamily}}
{\theorembodyfont{\rmfamily}\newtheorem{Mystat}{Statement}}
{\theorembodyfont{\rmfamily}}
\theoremheaderfont{\scshape}

\newcommand{\PreserveBackslash}[1]{\let\temp=\\#1\let\\=\temp}
\let\PBS=\PreserveBackslash

\title{On Application of the Local Search and the Genetic Algorithms Techniques to Some Combinatorial Optimization Problems}
\author{Anton Bondarenko \\
\texttt{anton.bondarenko@gmail.com}}
\date{April 2010} 
\begin{document}
\maketitle

%
%
%
%
%
%
%
\begin{abstract}
In this paper the approach to solving several combinatorial optimization problems using the local search and the genetic algorithm techniques is proposed. Initially this approach was developed in purpose to overcome some difficulties inhibiting the application of above mentioned techniques to the problems of the \emph{Questionnaire Theory}. But when the algorithms were developed it became clear that them could be successfully applied also to the \textsc{Minimum Set Cover}, the \textsc{0-1-Knapsack} and probably to other combinatorial optimization problems.
\end{abstract}

\textsc{Minimum Set Cover},
\textsc{Weighted Set Cover},
\textsc{0-1-Knapsack},
Local search,
Genetic algorithms

%
%
%
%
%
%
%
\section{Introduction}

%
%
%
\subsection{High-level overview of the proposed approach}
The \textsc{Optimal Binary Questionnaire} problem is $\mathcal{NP}$-hard \cite{ArCh:mftd}. In search of an efficient approximate algorithm several approaches were investigated and the special efforts were dedicated to the local search \cite{ArBo:tpem}, \cite{ArBo:tpait}.  

However all attempts to develop a neighborhood function for binary questionnaires have led only to a very limited success and a connected neighborhood has been found only for a tiny class of questionnaires having rather theoretical importance \cite{Bo:tpp}.

In this paper we propose to shift the research focus from the search within a set of questionnaires to the search within a set of functions of special kind. Such functions allow construction of the a questionnaire by consequential choice of questions for each subordinate problem starting from the root one. In this paper the set of \emph{root question selection functions} (RQSFs) effectively interconnected by the natural neighborhood relation is proposed.

After the implementation of the proposed algorithms it became clear that them could also be applied to some other combinatorial optimization problems, including \textsc{Minimum Set Cover}, \textsc{0-1-Knapsack} and probably to other ones. The background of this idea is given by the reductions of the mentioned combinatorial optimization problems to some questionnaire optimization problems used in the proofs of $\mathcal{NP}$-hardness and $\mathcal{NP}$-completeness of the questionnaire theory problems.

We expect that this approach could be successfully extended to many combinatorial optimization problem for which the efficient local search neighborhood and effective genetic operators haven't been found yet or do not exists at all.

In the remaining part of the paper we will show first how the proposed approach works for questionnaire optimization problems.  Then we will discuss briefly results of the laboratory testing of the developed algorithm. Having this done we will show how these algorithms could be applied to other combinatorial optimization problems. 

%
%
%
\subsection{The questionnaire theory basic definitions}
As it was stated above the mathematical model of binary questionnaire plays the central role in the presented approach and because this is not a widely known mathematical theory we will give here a brief introduction to it. From more details we suggest \cite{Pi:geq,PaSo:td,ArBo:pds}.

One of central tasks of the discrete search theory is the task of building of optimal in some sense conditional search strategy, i.e. the search strategy in which the choice of any test depends on outcomes of previously applied ones.  

One possible classification of discrete search problems is based on principles according to which test sets are formed. E.g. for the construction of \emph{binary tree}~\cite{H,Sob:gt} one can chose any possible subdivisions of a search area. And for \emph{binary search tree}~\cite{HT,GW} only tests preserving the linear order defined on the search area are allowed.

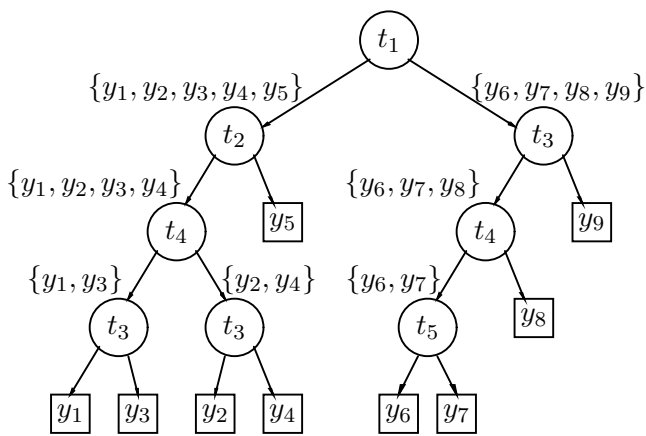
\begin{figure*}[htbp]
\begin{center} 
\setlength{\unitlength}{0.254mm}
\setlength{\unitlength}{0.254mm}
\begin{picture}(355,220)(0,-230)
       \allinethickness{0.254mm}\put(210,-25){\ellipse{30}{30}} 
       \put(195,-35){\makebox(30,20)[cc]{\shortstack{$t_1$}}} 
       \allinethickness{0.254mm}\put(130,-75){\ellipse{30}{30}} 
       \put(115,-85){\makebox(30,20)[cc]{\shortstack{$t_2$}}} 
       \allinethickness{0.254mm}\put(290,-75){\ellipse{30}{30}} 
       \put(275,-85){\makebox(30,20)[cc]{\shortstack{$t_3$}}} 
       \allinethickness{0.254mm}\put(260,-125){\ellipse{30}{30}} 
       \put(245,-135){\makebox(30,20)[cc]{\shortstack{$t_4$}}} 
       \allinethickness{0.254mm}\put(230,-175){\ellipse{30}{30}} 
       \put(215,-185){\makebox(30,20)[cc]{\shortstack{$t_5$}}} 
       \allinethickness{0.254mm}\put(70,-175){\ellipse{30}{30}} 
       \put(55,-185){\makebox(30,20)[cc]{\shortstack{$t_3$}}} 
       \allinethickness{0.254mm}\put(100,-125){\ellipse{30}{30}} 
       \put(85,-135){\makebox(30,20)[cc]{\shortstack{$t_4$}}} 
       \allinethickness{0.254mm}\path(145,-110)(165,-110)(165,-130)(145,-130)(145,-110) 
       \put(145,-130){\makebox(20,20)[cc]{\shortstack{$y_5$}}} 
       \allinethickness{0.254mm}\put(130,-175){\ellipse{30}{30}} 
       \put(115,-185){\makebox(30,20)[cc]{\shortstack{$t_3$}}} 
       \allinethickness{0.254mm}\path(145,-210)(165,-210)(165,-230)(145,-230)(145,-210) 
       \put(145,-230){\makebox(20,20)[cc]{\shortstack{$y_4$}}} 
       \allinethickness{0.254mm}\path(110,-210)(130,-210)(130,-230)(110,-230)(110,-210) 
       \put(110,-230){\makebox(20,20)[cc]{\shortstack{$y_2$}}} 
       \allinethickness{0.254mm}\path(70,-210)(90,-210)(90,-230)(70,-230)(70,-210) 
       \put(70,-230){\makebox(20,20)[cc]{\shortstack{$y_3$}}} 
       \allinethickness{0.254mm}\path(35,-210)(55,-210)(55,-230)(35,-230)(35,-210) 
       \put(35,-230){\makebox(20,20)[cc]{\shortstack{$y_1$}}} 
       \allinethickness{0.254mm}\path(140,-85)(150,-110)\special{sh 1}\path(150,-110)(148,-107)(149,-107)(149,-107)(150,-110) 
       \allinethickness{0.254mm}\path(140,-185)(150,-210)\special{sh 1}\path(150,-210)(148,-207)(149,-207)(149,-207)(150,-210) 
       \allinethickness{0.254mm}\path(270,-135)(280,-160)\special{sh 1}\path(280,-160)(278,-157)(279,-157)(279,-157)(280,-160) 
       \allinethickness{0.254mm}\path(300,-85)(310,-110)\special{sh 1}\path(310,-110)(308,-107)(309,-107)(309,-107)(310,-110) 
       \allinethickness{0.254mm}\path(120,-85)(105,-110)\special{sh 1}\path(105,-110)(107,-108)(106,-107)(106,-107)(105,-110) 
       \allinethickness{0.254mm}\path(90,-135)(75,-160)\special{sh 1}\path(75,-160)(77,-158)(76,-157)(76,-157)(75,-160) 
       \allinethickness{0.254mm}\path(250,-135)(235,-160)\special{sh 1}\path(235,-160)(237,-158)(236,-157)(236,-157)(235,-160) 
       \allinethickness{0.254mm}\path(280,-85)(265,-110)\special{sh 1}\path(265,-110)(267,-108)(266,-107)(266,-107)(265,-110) 
       \allinethickness{0.254mm}\path(60,-185)(45,-210)\special{sh 1}\path(45,-210)(47,-208)(46,-207)(46,-207)(45,-210) 
       \allinethickness{0.254mm}\path(75,-190)(80,-210)\special{sh 1}\path(80,-210)(79,-207)(79,-207)(80,-207)(80,-210) 
       \allinethickness{0.254mm}\path(125,-190)(120,-210)\special{sh 1}\path(120,-210)(121,-207)(120,-207)(120,-207)(120,-210) 
       \allinethickness{0.254mm}\path(205,-210)(225,-210)(225,-230)(205,-230)(205,-210) 
       \put(205,-230){\makebox(20,20)[cc]{\shortstack{$y_6$}}} 
       \allinethickness{0.254mm}\path(235,-210)(255,-210)(255,-230)(235,-230)(235,-210) 
       \put(235,-230){\makebox(20,20)[cc]{\shortstack{$y_7$}}} 
       \allinethickness{0.254mm}\path(235,-190)(245,-210)\special{sh 1}\path(245,-210)(242,-205)(243,-205)(244,-205)(245,-210) 
       \allinethickness{0.254mm}\path(225,-190)(215,-210)\special{sh 1}\path(215,-210)(218,-205)(217,-205)(216,-205)(215,-210) 
       \allinethickness{0.254mm}\path(200,-35)(145,-70)\special{sh 1}\path(145,-70)(148,-69)(147,-68)(147,-68)(145,-70) 
       \allinethickness{0.254mm}\path(220,-35)(275,-70)\special{sh 1}\path(275,-70)(272,-69)(272,-68)(273,-68)(275,-70) 
       \allinethickness{0.254mm}\path(275,-160)(295,-160)(295,-180)(275,-180)(275,-160) 
       \put(275,-180){\makebox(20,20)[cc]{\shortstack{$y_8$}}} 
       \allinethickness{0.254mm}\path(305,-110)(325,-110)(325,-130)(305,-130)(305,-110) 
       \put(305,-130){\makebox(20,20)[cc]{\shortstack{$y_9$}}} 
       \allinethickness{0.254mm}\path(110,-135)(125,-160)\special{sh 1}\path(125,-160)(123,-158)(123,-157)(124,-157)(125,-160) 
       \put(240,-60){\makebox(115,20)[cc]{\shortstack{$\{y_6, y_7, y_8, y_9\}$}}} 
       \put(40,-60){\makebox(140,20)[cc]{\shortstack{$\{y_1, y_2, y_3, y_4, y_5\}$}}} 
       \put(0,-110){\makebox(115,20)[cc]{\shortstack{$\{y_1, y_2, y_3, y_4\}$}}} 
       \put(15,-160){\makebox(65,20)[cc]{\shortstack{$\{y_1, y_3\}$}}} 
       \put(115,-160){\makebox(65,20)[cc]{\shortstack{$\{y_2, y_4\}$}}} 
       \put(180,-160){\makebox(65,20)[cc]{\shortstack{$\{y_6, y_7\}$}}} 
       \put(175,-110){\makebox(95,20)[cc]{\shortstack{$\{y_6, y_7, y_8\}$}}} 
\end{picture}
\caption{Example of arborescence for the task in the table~\ref{tab:1}}
\label{fig:3_1}
\end{center} 
\end{figure*}

Both \textsc{Optimal Binary Tree} and \textsc{Optimal Binary Search Tree} problems can be generalized within this classification in the following natural way.  Let's consider the problem of construction of conditional search strategy from a limited set of tests given by an explicit enumeration. The example of such problem is presented in the table~\ref{tab:1} and the one possible search strategy is given on the figure~\ref{fig:3_1}. Problems of this type are subject of the \emph{Questionnaire Theory} (QT) \cite{Pi:geq,PaSo:td,ArCh:obq}. 

\begin{table}[ht]
\begin{center}
\begin{tabular}{|c||c|} \hline
$t$ & \textsf{Outcomes} \\ \hline\hline
1 & 0:$\{y_1, y_2, y_3, y_4, y_5\}$, 1:$\{y_6, y_7, y_8, y_9\}$ \\ \hline
2 & 0:$\{y_1, y_2, y_3, y_4\}$, 1:$\{y_5, y_6, y_7, y_8, y_9\}$ \\ \hline
3 & 0:$\{y_1, y_2, y_5, y_6, y_7, y_8\}$, 1:$\{y_3, y_4, y_9\}$ \\ \hline
4 & 0:$\{y_1, y_3, y_5, y_6, y_7, y_9\}$, 1:$\{y_2, y_4, y_8\}$ \\ \hline
5 & 0:$\{y_1, y_2, y_3, y_4, y_5, y_6, y_8, y_9\}$, 1:$\{y_7\}$ \\ \hline
\end{tabular}
\end{center}
\caption{Example problem $A_1$\label{tab:1}}
\end{table}

\begin{figure*}[htbp]
\begin{center} 
\setlength{\unitlength}{0.254mm}
\setlength{\unitlength}{0.254mm}
\begin{picture}(440,340)(35,-350)
       \allinethickness{0.254mm}\put(380,-125){\ellipse{30}{30}} 
       \allinethickness{0.254mm}\put(90,-125){\ellipse{30}{30}} 
       \allinethickness{0.254mm}\put(360,-295){\ellipse{30}{30}} 
       \allinethickness{0.254mm}\put(430,-175){\ellipse{30}{30}} 
       \allinethickness{0.254mm}\put(360,-175){\ellipse{30}{30}} 
       \allinethickness{0.254mm}\put(230,-175){\ellipse{30}{30}} 
       \allinethickness{0.254mm}\put(150,-125){\ellipse{30}{30}} 
       \allinethickness{0.254mm}\put(300,-175){\ellipse{30}{30}} 
       \allinethickness{0.254mm}\put(330,-75){\ellipse{30}{30}} 
       \allinethickness{0.254mm}\put(120,-75){\ellipse{30}{30}} 
       \allinethickness{0.254mm}\put(280,-125){\ellipse{30}{30}} 
       \allinethickness{0.254mm}\put(225,-25){\ellipse{30}{30}} 
       \allinethickness{0.254mm}\path(210,-30)(135,-70)\special{sh 1}\path(135,-70)(140,-69)(140,-68)(140,-67)(135,-70) 
       \allinethickness{0.254mm}\path(240,-30)(315,-70)\special{sh 1}\path(315,-70)(310,-69)(310,-68)(310,-67)(315,-70) 
       \put(210,-35){\makebox(30,20)[cc]{\shortstack{$t_4$}}} 
       \put(105,-85){\makebox(30,20)[cc]{\shortstack{$t_1$}}} 
       \put(315,-85){\makebox(30,20)[cc]{\shortstack{$t_3$}}} 
       \allinethickness{0.254mm}\path(110,-85)(95,-110)\special{sh 1}\path(95,-110)(99,-106)(98,-105)(97,-104)(95,-110) 
       \allinethickness{0.254mm}\path(320,-85)(290,-115)\special{sh 1}\path(290,-115)(295,-112)(294,-111)(293,-110)(290,-115) 
       \put(75,-135){\makebox(30,20)[cc]{\shortstack{$t_5$}}} 
       \put(135,-135){\makebox(30,20)[cc]{\shortstack{$t_3$}}} 
       \put(365,-135){\makebox(30,20)[cc]{\shortstack{$t_1$}}} 
       \put(215,-185){\makebox(30,20)[cc]{\shortstack{$t_1$}}} 
       \allinethickness{0.254mm}\path(270,-135)(240,-165)\special{sh 1}\path(240,-165)(245,-162)(244,-161)(243,-160)(240,-165) 
       \allinethickness{0.254mm}\path(140,-135)(125,-160)\special{sh 1}\path(125,-160)(129,-156)(128,-155)(127,-154)(125,-160) 
       \allinethickness{0.254mm}\path(375,-140)(365,-160)\special{sh 1}\path(365,-160)(368,-155)(367,-155)(366,-155)(365,-160) 
       \put(285,-185){\makebox(30,20)[cc]{\shortstack{$t_6$}}} 
       \put(345,-185){\makebox(30,20)[cc]{\shortstack{$t_5$}}} 
       \put(415,-185){\makebox(30,20)[cc]{\shortstack{$t_6$}}} 
       \put(265,-135){\makebox(30,20)[cc]{\shortstack{$t_5$}}} 
       \allinethickness{0.254mm}\path(285,-140)(295,-160)\special{sh 1}\path(295,-160)(292,-155)(293,-155)(294,-155)(295,-160) 
       \allinethickness{0.254mm}\path(100,-135)(115,-160)\special{sh 1}\path(115,-160)(111,-156)(112,-155)(113,-154)(115,-160) 
       \put(345,-305){\makebox(30,20)[cc]{\shortstack{$t_1$}}} 
       \allinethickness{0.254mm}\path(160,-135)(175,-160)\special{sh 1}\path(175,-160)(171,-156)(172,-155)(173,-154)(175,-160) 
       \allinethickness{0.254mm}\path(80,-135)(65,-160)\special{sh 1}\path(65,-160)(69,-156)(68,-155)(67,-154)(65,-160) 
       \allinethickness{0.254mm}\path(50,-160)(70,-160)(70,-180)(50,-180)(50,-160) 
       \allinethickness{0.254mm}\path(170,-160)(190,-160)(190,-180)(170,-180)(170,-160) 
       \allinethickness{0.254mm}\path(110,-160)(130,-160)(130,-180)(110,-180)(110,-160) 
       \allinethickness{0.254mm}\path(340,-85)(370,-115)\special{sh 1}\path(370,-115)(365,-112)(366,-111)(367,-110)(370,-115) 
       \allinethickness{0.254mm}\path(130,-85)(145,-110)\special{sh 1}\path(145,-110)(141,-106)(142,-105)(143,-104)(145,-110) 
       \allinethickness{0.254mm}\path(390,-135)(420,-165)\special{sh 1}\path(420,-165)(415,-162)(416,-161)(417,-160)(420,-165) 
       \put(50,-180){\makebox(20,20)[cc]{\shortstack{$y_3$}}} 
       \put(110,-180){\makebox(20,20)[cc]{\shortstack{$y_6$}}} 
       \put(170,-180){\makebox(20,20)[cc]{\shortstack{$y_9$}}} 
       \allinethickness{0.254mm}\path(220,-185)(205,-210)\special{sh 1}\path(205,-210)(209,-206)(208,-205)(207,-204)(205,-210) 
       \allinethickness{0.254mm}\path(190,-210)(210,-210)(210,-230)(190,-230)(190,-210) 
       \put(190,-230){\makebox(20,20)[cc]{\shortstack{$y_1$}}} 
       \allinethickness{0.254mm}\path(295,-190)(285,-210)\special{sh 1}\path(285,-210)(288,-205)(287,-205)(286,-205)(285,-210) 
       \allinethickness{0.254mm}\path(275,-210)(295,-210)(295,-230)(275,-230)(275,-210) 
       \put(275,-230){\makebox(20,20)[cc]{\shortstack{$y_5$}}} 
       \allinethickness{0.254mm}\path(355,-190)(345,-210)\special{sh 1}\path(345,-210)(348,-205)(347,-205)(346,-205)(345,-210) 
       \allinethickness{0.254mm}\path(335,-210)(355,-210)(355,-230)(335,-230)(335,-210) 
       \put(335,-230){\makebox(20,20)[cc]{\shortstack{$y_8$}}} 
       \allinethickness{0.254mm}\path(420,-185)(405,-210)\special{sh 1}\path(405,-210)(409,-206)(408,-205)(407,-204)(405,-210) 
       \allinethickness{0.254mm}\path(390,-210)(410,-210)(410,-230)(390,-230)(390,-210) 
       \put(390,-230){\makebox(20,20)[cc]{\shortstack{$y_2$}}} 
       \allinethickness{0.254mm}\path(350,-305)(335,-330)\special{sh 1}\path(335,-330)(339,-326)(338,-325)(337,-324)(335,-330) 
       \allinethickness{0.254mm}\path(320,-330)(340,-330)(340,-350)(320,-350)(320,-330) 
       \put(320,-350){\makebox(20,20)[cc]{\shortstack{$y_7$}}} 
       \allinethickness{0.254mm}\path(370,-305)(385,-330)\special{sh 1}\path(385,-330)(381,-326)(382,-325)(383,-324)(385,-330) 
       \allinethickness{0.254mm}\path(380,-330)(400,-330)(400,-350)(380,-350)(380,-330) 
       \put(380,-350){\makebox(20,20)[cc]{\shortstack{$y_4$}}} 
       \allinethickness{0.254mm}\path(305,-190)(355,-280)\special{sh 1}\path(355,-280)(352,-275)(353,-275)(354,-275)(355,-280) 
       \allinethickness{0.254mm}\path(360,-190)(360,-280)\special{sh 1}\path(360,-280)(362,-274)(360,-274)(358,-274)(360,-280) 
       \allinethickness{0.254mm}\path(425,-190)(390,-330)\special{sh 1}\path(390,-330)(392,-325)(391,-325)(390,-325)(390,-330) 
       \allinethickness{0.254mm}\path(235,-190)(325,-330)\special{sh 1}\path(325,-330)(321,-326)(322,-325)(323,-324)(325,-330) 
       \put(95,-50){\makebox(90,20)[cc]{\shortstack{$\{y_3, y_6, y_9\}$}}} 
       \put(35,-110){\makebox(70,20)[cc]{\shortstack{$\{y_3, y_6\}$}}} 
       \put(135,-110){\makebox(70,20)[cc]{\shortstack{$\{y_6, y_9\}$}}} 
       \put(260,-50){\makebox(185,20)[cc]{\shortstack{$\{y_1, y_2, y_4, y_5, y_7, y_8\}$}}} 
       \put(350,-110){\makebox(115,20)[cc]{\shortstack{$\{y_2, y_4, y_7, y_8\}$}}} 
       \put(405,-160){\makebox(70,20)[cc]{\shortstack{$\{y_2, y_4\}$}}} 
       \put(185,-160){\makebox(70,20)[cc]{\shortstack{$\{y_1, y_7\}$}}} 
       \put(270,-255){\makebox(70,20)[cc]{\shortstack{$\{y_4, y_7\}$}}} 
       \put(350,-255){\makebox(70,20)[cc]{\shortstack{$\{y_4, y_7\}$}}} 
       \put(290,-145){\makebox(95,15)[cc]{\shortstack{$\{y_4, y_7, y_8\}$}}} 
       \put(285,-160){\makebox(85,15)[cc]{\shortstack{$\{y_4, y_5, y_7\}$}}} 
       \put(200,-110){\makebox(105,20)[cc]{\shortstack{$\{y_1, y_4, y_5, y_7\}$}}} 
\end{picture}
\caption{Example of latticoid questionnaire for the task in the table~\ref{tab:2}}
\label{fig:3_2}
\end{center} 
\end{figure*}
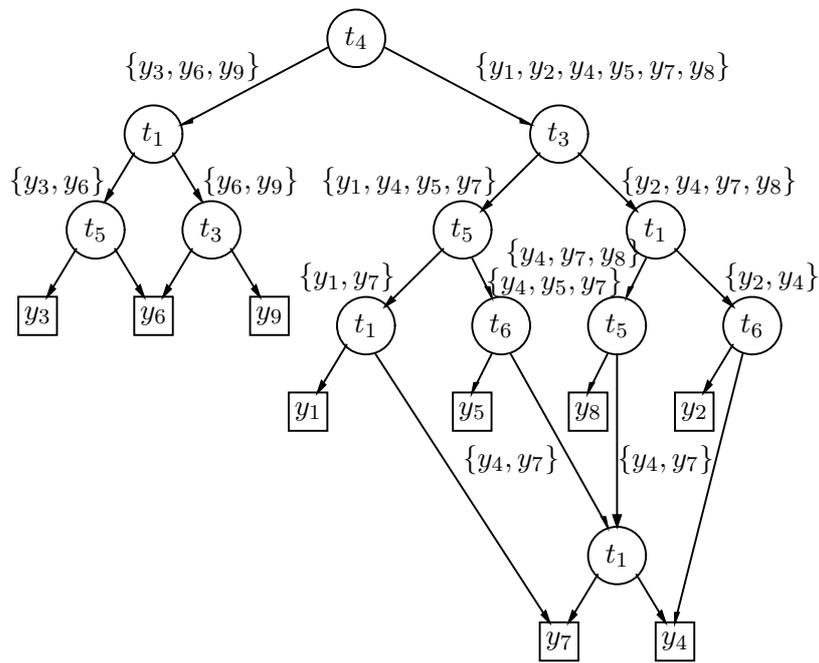

Two types of tests (called \emph{questions}) are considered in the theory of questionnaires. Each question of the first type defines subdivision of a search area into independent classes. Outcomes of each question of the second type can have nonempty intersections and covers the search area. In the first case the questionnaire can be represented by a rooted tree. Questionnaires of the second type are represented by acyclic graphs with a single source vertex. Picard~\cite{Pi:geq} called questionnaires of the first type \emph{arborescences} and questionnaires of the second type \emph{latticoids}. In this paper we will consider only arborescent questionnaires. Example of an arborescent quesitionnaire is given in the table~\ref{tab:1} and on the figure~\ref{fig:3_1}, example of an latticoid questionnaire is given in the table~\ref{tab:2} and on the figure~\ref{fig:3_2}.

Application of the question within the questionnaire breaks the problem table into several tables, one per outcome of the question. Thus for binary question there will be 2 'sub-tables'. Each derived table is formed as a column subset of the basic problem table with the outcome number in the row represented the 'asked' question equal to the outcome number. Thus '$0$-subproblem' table of the question $t_i$ will contain all columns with $'0'$ in the $i$-th position.

Questions containing a single outcome are called \emph{senseless}. In particular senseless questions can be found in problem tables obtained after application of some question. Senseless questions are removed from problem tables. 

\begin{table}[ht]
\begin{center}
\begin{tabular}{|c||c|} \hline
$t$ & \textsf{Outcomes} \\ \hline\hline
1 & 0:$\{y_1, y_3, y_4, y_5, y_6, y_7, y_8\}$, 1:$\{y_2, y_4, y_5, y_6, y_9\}$ \\ \hline
2 & 0:$\{y_1, y_2, y_3, y_4, y_5, y_6, y_9\}$, 1:$\{y_3, y_4, y_7, y_8\}$ \\ \hline
3 & 0:$\{y_1, y_3, y_4, y_5, y_6, y_7\}$, 1:$\{y_2, y_3, y_4, y_7, y_8, y_9\}$ \\ \hline
4 & 0:$\{y_3, y_6, y_9\}$, 1:$\{y_1, y_2, y_4, y_5, y_7, y_8\}$ \\ \hline
5 & 0:$\{y_1, y_2, y_3, y_7, y_8, y_9\}$, 1:$\{y_4, y_5, y_6, y_7, y_8, y_9\}$ \\ \hline
6 & 0:$\{y_1, y_2, y_3, y_5, y_6, y_8, y_9\}$, 1:$\{y_1, y_2, y_3, y_4, y_6, y_7, y_9\}$ \\ \hline
7 & 0:$\{y_2, y_3, y_6, y_7, y_8, y_9\}$, 1:$\{y_1, y_2, y_3, y_4, y_5, y_6, y_8, y_9\}$ \\ \hline
\end{tabular}
\end{center}
\caption{$A_2$\label{tab:2}}
\end{table}

A number of possible outcomes of the question called the \emph{question base}. We will consider in this paper only \emph{binary} questions, i.e. questions of the base 2. Questionnaires built from binary questions are respectively called \emph{binary}.

Search area $Y$ is considered traditionally as a set of independent events $y_j$ with a given discrete distribution $p(y_j)=p_j$. The convenient representation of set of questions $T$ is given by a table where each row represents one question and the number on intersection of row $i$ and column $j$ represents the outcome of question $i$ which event $y_j$ belongs to. Table~\ref{tab:3} contains the same set of questions as the table~\ref{tab:1} does but represented in the manner we just described.

\begin{table}[ht]
\begin{center}
\newcolumntype{M}{>{\small\PBS\centering}p{9mm}}
\begin{tabular}{|c||c||M|M|M|M|M|M|M|M|M|} \hline
$t$ & $c(t)$ : $y$ &  1 & 2 & 3 & 4 & 5 & 6 & 7 & 8 & 9  \\ \hline\hline
1 & 3,00 & 0 & 0 & 0 & 0 & 0 & 1 & 1 & 1 & 1 \\ \hline
2 & 7,00 & 0 & 0 & 0 & 0 & 1 & 1 & 1 & 1 & 1 \\ \hline
3 & 4,00 & 0 & 0 & 1 & 1 & 0 & 0 & 0 & 0 & 1  \\ \hline
4 & 5,00 & 0 & 1 & 0 & 1 & 0 & 0 & 0 & 1 & 0 \\ \hline
5 & 6,00 & 0 & 0 & 0 & 0 & 0 & 0 & 1 & 0 & 0 \\ \hline\hline
& $p(y)$ & 0,05 & 0,10 & 0,05 & 0,30 & 0,20 & 0,05 & 0,05 & 0,15 & 0,05 \\ \hline
\end{tabular}
\end{center}
\caption{$A_3$\label{tab:3}}
\end{table}

Another aspect the questionnaire theory extends the traditional discrete search models in is the cost of individual tests. While for optimization of binary trees and binary search trees we traditionally take 1 as a cost of each test, the questionnaire theory allows to define the \emph{cost} function on the set of questions $c(t_i)=c_i \in\mathbb{R}$. The sum of costs of questions applied in the current questionnaire to identify some particular event $y_i$ is called \emph{cost of identification} of $y_i$. The mean value of cost of identification of events from the search area $Y$ for the given questionnaire $Q$ is called \emph{cost} of questionnaire $C(Q)=\sum_{y_i \in Y}c(y_i)$ where $c(y_i)$ is the sum of cost of questions applied to identify $y_i$ in $Q$.

E.g. the cost of questionnaire on picture~\ref{fig:3_1} is:
\begin{equation}
\begin{split}
C(Q) = &(c_1+c_2+c_3+c_4)(y_1+p(y_2)+p(y_3)+p(y_4))+ \\
       &(c_1+c_2)p(y_5)+ \\
       &(c_1+c_3+c_4+c_5)(p(y_6)+p(y_7))+ \\
       &(c_1+c_3+c_4)p(y_8)+ \\
       &(c_1+c_3)p(y_9)=18,45 \label{eq:1}
\end{split}
\end{equation}

Multiply questionnaires of different cost can be constructed for each individual problem and in this paper we will consider the problem of building of optimal questionnaire in the sense of the defined cost function.

We call the task of constructing an optimal not weighted (with all questions' cost equal 1) questionnaire \textsc{Optimal Questionnaire} (OQ) or \textsc{Optimal Binary Questionnaire} (OBQ) for the binary case. Weighted versions of these tasks will be called, respectively \textsc{Optimal Weighted Questionnaire} (OWQ) and \textsc{Optimal Weighted Binary Questionnaire} (OWBQ). All these problems will be called the \emph{problems of the theory of questionnaires}.

We call an individual problem of the theory of questionnaires logically complete if any pair of events is separated at least by one question. Otherwise obviously it is impossible to construct a questionnaire that identifies all the events.

Obviously, for logical completeness any questionnaire theory problem it is necessary and sufficient that any pair of columns in its table of questions differs in at least one position. Further in this paper we always assume the logical completeness of considered problems. 

%
%
%
\subsection{Complexity and approximability}
\begin{Mystat} 
OBQ is $\mathcal{NP}$-hard.
\end{Mystat}\label{st:obq_np}
{\scshape Proof}. We will present the reduction from \textsc{Minimum Set Cover} (MSC) \cite{GJ} to OBQ. Let $M$ be the individual MSC with the universe $U$ and the family $S$ of subsets of $U$ represented as a binary $n \times k$-table, where $n=|U|$, $k=|S|$ and the intersection of column $i$ and row $j$ contain $'1'$ if $u_j \in S_i$ and $'0'$ otherwise. We assume that the described table doesn't contain similar columns. Otherwise we can combine each subset of similar columns together without loss of generality.

We will construct the table representation of the derived individual OBQ by adding to the table of the individual MSC a column $y_0$ consisting of zeroes. Since the added column will differ from any column in the original MSC table and this original table also doesn't contain similar columns as we discussed the derived OBQ table will be logically complete and thus will allow a construction of a complete questionnaire.

Let assign probabilities to the events of the derived OBQ table in the following way:
\begin{equation}
p_i = 
   \begin{cases}
      \varepsilon ,\varepsilon\,n\,(k+1)<1,& \text{if i $\neq 0$} \\
      1-n\varepsilon,& \text{if $i=0$}
   \end{cases}
\end{equation}\label{ea:prpr}

We will show now that if $Q$ is the optimal questionnaire for the derived OBQ, then the set of subsets of $U$ corresponding to the questions constituting the branch of the $Q$ spanning the root and the event $y_0$ will represent the minimal cover for the original MSC.

It is straightforward to show that the constructed set represents a cover for the original individual MSC problem. Now we will show that it is the minimal cover.

Let assume the opposite, that is there exists the cover $T'$ for the original MSC problem, such that $|T'|<|T|$. Then it is possible to build the questionnaire $Q'$ which will identify event $y_0$ using $|T'|$ questions corresponding to the elements of $T'$. Indeed since the elements of $T'$ cover all the elements of $U$, for each event $y_i \in \{y_1,...,y_n\}$ there exists an element of $S$ belonging to the $T'$ which contains '1' in the $i$-th positions and therefore which distinguishes $y_i$ from $y_0$. The rest of the questionnaire is not important, it is important only that we can build a complete questionnaire since the problem table logically complete.

The cost of the $Q'$ will satisfy the following inequality:
$C(Q') \le |T'|\,p_0 + nke$

The cost of $Q$ is $C(Q)=|T|\,p_0+le$ where $l$ - is the sum of lengths of branches spanning the root of $Q$ with the events $e_1,...,e_n$. According to our assumption $|T'|<|T|$ and therefore $|T'|-|T|\geq1$ and according to the rule \ref{ea:prpr} $nk(\varepsilon + 1)<1$ and finally: 
$C(Q)-C(Q') = p_0(|T'|-|T|)+l\varepsilon-nk\varepsilon>p_0-nk\varepsilon=1-n\varepsilon-nk\varepsilon=1-nk(\varepsilon+1)>0$ and thus $Q$ is not optimal. Contradiction.$\square$

\begin{Mystat}  
OQ, OWQ, OWBQ are $\mathcal{NP}$-hard
\end{Mystat}\label{st:q_np}
{\scshape Proof}. OQ, OWQ, OWBQ are all generalizations of OBQ.$\square$

Since all significant problems of the questionnaire theory are $\mathcal{NP}$-hard the task of development of efficient approximate algorithms becomes very important. But before discussing the proposed algorithm we will check to what extent the OBQ and OWBQ are approximable. It will let us set the proper goals regarding the quality of the developed algorithms.

Feige \cite{F} showed that MSC cannot be approximated in polynomial time within a factor of $(1-\mathcal{O}(1))\ln n$ unless $\mathcal{NP}$ has quasi-polynomial time algorithm, i.e. unless $\mathcal{NP} \subset\textsc{Dtime}(n^{\log\log n})$. Raz and Safra \cite{RS} established a lower bond of $c\ln n$, where $c$ is a constant under the weaker assumption $\mathcal{P}\neq\mathcal{NP}$. 

Since we can choose $\varepsilon$ as small as necessary and thus make the difference between the size of optimal cover and the cost of the corresponding questionnaire as small as we wish, the results about the MSC inapproximability will also apply to OBQ and OWBQ. In the rest of this paper we will show how a local search and genetic algorithms can be applied to problems of the questionnaire theory and to some extent we will generalize the proposed approach for the \textsc{Minimum Set Cover} and \textsc{0-1-Knapsack} problems.

Significant efforts were spent on attempts to apply the local search approach to the OWBQ \cite{ArBo:tpait, ArBo:tpem}. Unfortunately the described in these papers algorithm is only applicable to the very specific case. In this case OWBQ forms a matroid and as a result the exact solution could be obtained through using of greedy algorithm \cite{Bo:tpp}. These difficulties is a consequence of a relatively high internal complexity of the binary questionnaire model which makes it actually impossible to develop efficient neighborhood operators for the local search method as well as to develop a correct and efficient crossover and mutation operators for the genetic algorithm implementation.

%
%
%
%
%
%
%
\section{Local Search}

%
%
%
\subsection{Simple greedy strategies}
We will start the construction of the proposed algorithm from the investigation of characteristics of simple greedy strategies. Several elementary greedy \emph{root question selection functions} (RQSF) are represented in the table \ref{tab:4}. These functions allow the construction of a questionnaire in the top-down manner by consecutive choise of the root question for the produced on the previous steps subordinate problems. 

The numerous laboratory test has shown that in the most part of cases the function No.4 demonstrates the best performance among all. However in the same time for some cases other functions can be more efficient.

%
%
%
\subsection{Composite strategy}
As it was mentioned different RQSFs can deliver better solutions for different individual OWBQ problems. So we can expect that this property will hold also for any set of subordinate problems of the given individual OWBQ problem.

Keeping this property in mind we will split the set of all individual OWBQ problems into a finite set of classes and assign to each class some type of RQSF. Such composite RQSFs form a space with natural neighborhoud function based on replacement of elementary RQSF assigned to different classes of OWBQ problems in the composite RQSF.

We will extend the basic set of RQSFs in the table \ref{tab:4} with some artificial functions which although cannot be considered as optimizing strategies themselves but which are very useful for overcoming of local extremums. We will discuss these functions with more details in the section \ref{sec:res}.

\renewcommand\arraystretch{2.3}
\begin{table}[ht]
\begin{center}
\begin{tabular}{|c||c|l|} \hline
\textsf{No.} & $f$ & \textsf{Comments} \\ \hline\hline
1 & $\arg \min c(t)$ & Question cost \\ \hline
2 & $\arg \max \Delta\,H$ & \multicolumn{1}{m{60mm}|}{Maximal decrease of entropy} \\ \hline
3 & $\arg \max \frac{\Delta\,H}{c(t)}$ & Maximal decrease of entropy to cost \\ \hline
4 & \multicolumn{1}{m{65mm}|}{$\arg \min (\frac{c(t)}{p_0(t)}+\frac{c(t)}{p_1(t)})$, where $p_s(t)$ - is the sum of probabilities of outcome $s$ for $t$} & Question preference function \\  \hline
\end{tabular}
\end{center}
\caption{Greedy functions\label{tab:4}}
\end{table}
\renewcommand\arraystretch{1}
    
%
%
%
\subsection{Decomposition of the space of subordinate problems}
For the partition of the set of individual OWBQ problems into classes, we will choose some characteristic function that maps a set of individual problems into $\mathbb{R}$. Table\ref{tab:5} contains the potential candidates for the role of such characteristic function. 

\renewcommand\arraystretch{2.3}
\begin{table}[ht]
\begin{center}
\begin{tabular}{|c||c|c|c|l|} \hline
\textsf{No.} & $f$ &  $f_{max}$ & $f_{min}$ & \textsf{Comments} \\ \hline\hline
1 & $H=-\sum_{i=0}^n\,p_i\,log_2\,p_i$ & $log_2\,n$ & $0$ & Entropy \\ \hline
2 & $\dfrac{n}{r}$ & $\dfrac{2^r}{r}$ & $\dfrac{r}{2^r}$ & Compactness \\ \hline
3 & \multicolumn{1}{m{55mm}|}{$H_c=-\sum_{j=0}^r\,c'_j\,log_2\,c'_j$, where $c'_j$ - is the 'discrete' cost 'distribution', e.g. $\sum_{j=0}^r\,c'_j=1$}
& $log_2\,r$ & $0$ & \multicolumn{1}{m{45mm}|}{Entropy of cost 'distribution'}  \\ \hline
\end{tabular}
\end{center}
\caption{Characteristic functions\label{tab:5}}
\end{table}
\renewcommand\arraystretch{1}

Based on laboratory testing it was revealed that among all the functions presented in the table the entropy $H(T)$ allows to obtain the most uniform distribution of values for the subordinate problems of a given individual OWBQ problem in most cases and we shall use $H(T)$ in the proposed algorithm.

To split a set of individual OWBQ problems into classes, we need to break the range of the selected characteristic function into a finite number of intervals. Each of the selected intervals will induce a corresponding class of equivalence on the set of individual problems.

An obvious approach is to choose a certain number of intervals of equal length to be determined depending on the size of the problem. However, this approach leaves some room for improvement.

Uneven distribution of values of the characteristic function between the intervals can lead to the situation when the part of the induced classes will contain several subordinate problems each, and the other ones can remain empty. As a result, the flexibility of combined function will decline.

An attempt to compensate for this shortcoming by increasing the number of intervals will result in increased complexity of the algorithm. The solution is to use a set of intervals of variable size, such that each subordinate problem corresponds to exactly one interval.

The number of subordinate problems of the given individual OWBQ problem is equal to the number of vertices of the arbitrary questionnaire of this problem, and thus is equal to $n-1$. We will choose the boundaries between the intervals in the middle between adjacent pairs of values of an ordered sequence of values of the characteristic function calculated for the set of subordinate problems defined by the current questionnaire. In other words, the system of intervals will be dynamic and will depend on the current questionnaire, or to be more precise, on the set of subordinate problems defined by the current questionnaire. 

Obviously, the changes of the questionnaire, carried out at each step of local search, will also affect the set of subordinate problems, and as a consequence, at each step the system of intervals will require adjustment.

%
%
%
\subsection{The algorithm}
To represent the composite RQSF we will use the table containing two rows. The first row of this table will contain the (upper) boundaries of intervals of values of the characteristic function that is used to split the set of subordinate problems of the resolved OWBQ problem into subsets. The second row will contain the type values of the elementary RQSFs assigned for appropriate intervals. 

As it was described in the previous section we will use the $n-1$ intervals with variable boundaries chosen midway between adjacent pairs of elements of an ordered sequence of values of the characteristic function calculated for the set of subordinate problems corresponding to the internal vertexes (to the questions) of the current questionnaire.

On each iteration of local search the algorithm produces a neighborhood of the current solution sequentially changing the type of elementary root question selection function assigned to each interval. The system of intervals of the characteristic function is remaining unchanged until the moment of choosing the cheapest neighbor.

If the found solution is cheaper than current, it is designated as current one and the system of intervals is updated based on the set of subordinate problems generated by the questionnaire, constructed using the updated combined RQSF. Elementary RQSF type values are assigned to the newly created intervals in the manner preserving the RQSF types applied to the subordinate problems of the resolved individual OWBQ problem before the update of the interval system. Obviously, in this approach after the update intervals of the characteristic function, the current composite RQSF will generate the same questionnaire as before the update.

\begin{algo}
\large
\caption{{\large Local search for OWBQ}\label{algo:ls}}
\begin{verbatim}
/*
   F - combined RQSF 
   f = F[i] - RQSF i 
   Q = F(T) - questionnaire which is outcome of 
       the function F for individual problem T 
   G - Set of elementary RQSFs
*/

<Choose the initial combined function Fnew>
do{
   Fcurrent = Fnew;
   for(int i = 0; i < |Fc|; i++)
      foreach(f' : G)
         if(f' != Fcurrent[i]){
            F' = Fcurrent;
            F'[i] = f';
            if(C(Fnew(T)) < C(F'(T)))
               Fnew = Fcurrent;
         }
} until (C(Fnew(T)) < C(Fcurrent(T)))
     
\end{verbatim}
\normalsize
\end{algo}

%
%
%
\subsection{Analysis of the test results}\label{sec:res}
Results of testing of the algorithm \ref{algo:ls} represented in the table \ref{tab:7} let conclude that the algorithm is quite efficient and produce mostly better or similar solutions then known approximation algorithms do.

However in a substantial number of cases, the proposed algorithm did not improve the results of elementary methods. In addition to the above results about $\mathcal{NP}$-hardness and nonapproximability of OWBQ/OBQ there is also another reason for the complexity of developing high-quality approximation algorithms for these problems.

The space of solutions of a typical OWBQ problem contains a significant number of local extremums. This property of the questionnaire theory problems is a consequence of the tree structure of the questionnaire and the properties of the cost function allowing independent existence of multiple local extremums for different subtrees of a questionnaire both on independent branches and combined hierarchically.

It should be noted that the neighborhood function used in the proposed algorithm can link entirely different questionnaires to each other. For example, if the elementary function has been changed for the interval which contains the root problem, then beginning with the change of the root question, the construction of the questionnaire goes completely differently way. However, we still endeavored to make our algorithm more resistant to local extremums.

To achieve this we will extend the neighborhood by expanding the set of RQSFs with some special \textit{'dumb'} functions $F_k$, returning the constant question number $k$ each. In fact, the newly added functions will not be exactly the constant because with a decrease in the number of available questions during the gradual construction of a questionnaire part of questions become senseless and are removed from the problem table, so we need to ensure that any 'dumb' function returns values not exceeding the number of questions in the current problem table.

To achieve this the 'dumb' functions will return a value of $k\mod n$. The use of 'dumb' functions will, in fact, let the algorithm to do a step aside at each step thus trying to avoid a possible local extremum. The \textsf{Mixed} column of the table \ref{tab:7} presents the results of the algorithm \ref{algo:ls} tests with the extended neighborhood. 

Additional evidence of the justification for the inclusion of the discussed above 'dumb' functions are the results of testing of algorithm \ref{algo:ls} with 'dumb' functions only (see table \ref{tab:7} - column \textsf{'Dumb'}). Despite the fact that the lack of greedy functions the method is slightly worse but nevertheless is quite effective. 

%
%
%
%
%
%
%
\section{Genetic algorithms}
We will use tradition GA terminology, see eg. \cite{Ho, Be1, Be2}

%
%
%
\subsection{Representation of individuals}
Let us consider how the proposed approach can be used to develop a genetic algorithm (e.g. see \cite{Ho, Be1, Be2}) for the OWBQ. The coding of solutions in the form of a linear chain significantly simplifies the development of genetic operators. However, solution model that we used in the algorithm \ref{algo:ls}, has more complex structure. It includes the partition of the set of individual problems into classes induced by the set of intervals of some characteristic function together with the mapping reflecting these intervals to the set of subordinate problems defined by the questionnaire of this solution. 

Since we use intervals with variable boundaries, a simple substitution of some type value of elementary functions from one solution to another has little meaning since the function, type of which will be transferred, can be applied to the individual problems in the range of the characteristic function different than in the solution, from which it was borrowed.

Therefore, in order to simplify the genetic operators we are forced to fallback to the solution representation with intervals of equal length. However, in order to avoid too uneven distribution of sub-problem among intervals, we will increase the amount of intervals. During the laboratory tests the different approach of selection of the number of intervals were investigated, but the most effective were the values between $nr$ and $n^2r^2$.

%
%
%
\subsection{Genetic operators}
Since we switched to the simplified representation of individual OWBQ problems, which now is equivalent to a linear string of values, development of genetic operators becomes a trivial task.

To implement the \emph{crossover} operator it is enough to break two \emph{genotype} chains which we're going to cross over at a certain position and glue the pairs of obtained fragments from different chains together. During the laboratory tests some more complex operators were checked out including \emph{2-point} and \emph{uniform} crossover \cite{Be1}. However the real impact from these modifications was insufficient and we have chosen the simplest approach.

The mutation operator implemented is also fairly simple. The type value of an elementary root question selection function in a randomly selected position solution is replaced by another randomly selected type. 

However, due to the redundancy of the set of intervals, the replacement of a single gene has very little impact and it was decided to increase the number of genes that are changed within the single mutation. Different methods of choice of the number of genes which are subject to mutation have been tested and the value of $l/r$, where $l$ - the length of the genotype have been chosen as the most efficient one. 

%
%
%
\subsection{Choosing a strategy for the formation of generations}
There are different approaches to the formation of generations in genetic algorithms. In one approach, each new generation has the same size as the entire population, in another one it represents only some part of a population. Sometimes all members of a new generation are included entirely in a population displacing the least adapted members of previous generations, sometimes the competition between new and previous generations is implemented. We have chosen the option of several generations with a competitive incorporation of new members (see Algorithm \ref{algo:ga}).

Another important aspect of the strategy is the method of selection of individuals for emph{mating}. A cost of the questionnaire doesn't represent a good fitness function because the relative difference in cost of different questionnaires is quite small and doesn't provide enough advantage for cheaper solutions during selection. The reason for this is very small relative differences between the costs of various questionnaires. To ensure effective selection and to help to prevent premature convergence, we will use as a fitness function the questionnaire cost, scaled as follows: $\Phi(Q)=C(Q)-\min C(Q), Q\in G$, where $G$ - is the new generation. 

%
%
%
\subsection{Parameters of the algorithm}
The table \ref{tab:6} represents the key parameters of the  algorithm \ref{algo:ga}.
\begin{table}[ht]
\begin{center}
\begin{tabular}{|m{55mm}|m{80mm}|} \hline 
\textsf{Parameter name} & \textsf{Description} \\ \hline\hline
Mating Rate & Average number of matings per individual \\ \hline
Mutation Rate & Probability of mutation \\ \hline  
Length of genotype & Number of symbols in genotype \\ \hline
Number of generations without improvements & Parameter used in algorithm halt condition \\ \hline
Maximal total number of generations & Parameter used in algorithm halt condition \\ \hline
RQSF set & See table \ref{tab:4} and the 'dumb' functions\\ \hline
Characteristic function &  See table \ref{tab:5} \\ \hline
\end{tabular}
\end{center}
\caption{Parameters of genetic algorithm\label{tab:6}}
\end{table}

%
%
%

\begin{algo}
\large
\caption{{\large Genetic Algorithm for OWBQ}\label{algo:ga}}
\begin{verbatim}
/*
   improvement - difference between maximum fitness values 
   of two consecutive generations
*/
<prepare initial population>
iterationNo = 0;
generationNo = 0;
do {
   for(int i = 0; i < populationSize * matingRate; i++){
      <Choose male>;
      <Choose female>;
      <mate selected individuals>;
      <apply mutation to offsprings>; 
      <add offspins to population>;
   }
   while(|population| > populationSize)
      <remove least fit individual>;
   generationNo++;
   if (improvement == 0.0)
      iterationNo++;
   else
      iterationNo = 0;
} while (iterationNo < generationsWithoutImprovement 
      && generationNo < maxNumberOfGenerations);     
\end{verbatim}
\normalsize
\end{algo}

%
%
%
\subsection{Analysis of test results}
The main result of testing was the proof of the effectiveness of the Algorithm \ref{algo:ga}. For some part of the solutions the proposed algorithm provided better solutions (See table \ref{tab:7}) than ones obtained with the help of basic greedy methods and by the help of the algorithm \ref{algo:ls}. In many cases the algorithm founds solution of the same quality as one found with the basic greedy functions.

%
%
%
%
%
%
%
\section{Application of the developed method to other combinatorial optimization problems}
In this section we will discuss how the developed algorithm can be applied to the \textsc{Minimum Set Cover}, to the \textsc{Weighted Set Cover} and to the \textsc{0-1-Knapsack} problems.

%
%
%
\subsection{Minimum Set Cover and Weighted Set Cover}
Any individual MSC problem can be reduced to the OBQ using the method described in the proof of statement \ref{st:q_np}. As well as any \textsc{Weighted Set Cover} problem can be reduceded to OWBQ. The probability distribution of the obtained OBQ/OWBQ is quite specific and the entropy function loses its discriminative properties as a characteristic function. 

So we're switching to the Compactness function (see table \ref{tab:5}). For the \textsc{Weighted Set Cover} problem also the 'cost entropy' $H_c$ function can be used. It is also worthwhile to modify the set of RQSFs. Then algorithms \ref{algo:ls} and \ref{algo:ga} are applicable without any changes. 

%
%
%
\subsection{Combinatorial 0-1-knapsack}
We will show in this subsection how some modification of OWBQ can be used as a representation of \textsc{0-1-Knapsack} problem in purpose to make algorithms \ref{algo:ls} and \ref{algo:ga} applicable to these problems.

Consider a modification of OWBQ with a limited maximum length of the branches. This problem will be called the problem of the \textsc{Limited Depth Questionnaire} (LDQ) (see \cite{ArBo:ddv}). Let's get acquainted with this problem more.

It is obvious that in general the construction of the questionnaire which will fully identify the set of events impossible under the condition of restricted depth. As a result of this limitation and due to the properties of the considered problem the notion of the cost of the questionnaire as a criterion of optimality becomes meaningless. Therefore, we propose a criterion that would reflect the degree of identification of the set $L$ of events by the measured questionnaire.

We assume that each element of the search area $ L $ is assigned the \emph{weight} function $d(y_i)$, satisfying  the axioms of measure:

\begin{equation}
\begin{split}
&\forall L_1\subseteq L : d(L_1) \geq 1 \\
&\forall L_1\subset L : d(L_1)=0 \Rightarrow L_1 = \oslash \\
&\forall L_1, L_2 \subseteq L : d(L_1 \bigcup L_2) \leq d(L_1) + D(L_2) \\
&\forall L_1, L_2 \subseteq L : d(L_1 \bigcup L_2) = d(L_1) + d(L_2) \Rightarrow L_1 \bigcap L_2 = \oslash \\
&\forall L_1, L_2 \subset L : L_1 \subset L_2 \Rightarrow d(L_1) \le d(L_2) \label{eq:2} 
\end{split}
\end{equation}

We shall call $d(L^*)$ the \emph{size of set of events} $L*$. Let $d(L)=1$. We will consider the case when all elements of the search area have the same size: $\forall y_i:d(y_i)=1/n$. This approach reflects the situation when all events in $L$ have equal importance from the identification perspective.

A quantitative characterization of the degree of identification system $L$ with respect to its partition into subsets $L_1,...,L_k$ is given by:

\begin{equation}
D(L_1,...,L_k)=M(L_1,...,L_k)=\sum_{i=1}^kd(L_i)\sum_{y_j\in L_i}p(y_j) \label{eq:3}
\end{equation}

Obviously the less the value of $D$ the more the overall \emph{depth of identification} is. Thus, we will strive to minimize the average size of the partition produced by the questionnaire under condition that the summary cost of questions asked along any branch of the questionnaire should not exceed some specified value $c^*$.

\begin{Mystat} 
LDQ is $\mathcal{NP}$-complete.
\end{Mystat}\label{st:ldq_np}
{\scshape Proof}. Let $I$ - the individual \textsc{0-1-Knapsack} problem. We will form the corresponding LDQ as follows. For each element of $e_i \in I$ we will include the question $t_i$, which is a single-event check, i.e.:

\begin{equation}
\forall t_i : |L_s(t_i)|=1\,\&\,L_{\bar s}(t_i)\mid=n-1 \label{eq:4}
\end{equation}

We set the cost $c(t_i)=d(e_i)$, where $d(e_i)$ - is the weight of element $e_i$ in the reduced individual \textsc{0-1-Knapsack} problem $I$. Also we will put the probabilities of all events equal to each other. Suppose also $c^*=d^*$, where $d^*$ - the knapsack size in the problem $I$.

Obviously the optimal in the sense of criterion \ref{eq:3} questionnaire for the derived individual LDQ will correspond to the optimal packing of knapsack in the problem $I$.$\square$

Now, as in the case of the covering problems, we can apply algorithms \ref{algo:ls} and \ref{algo:ga} after replacing the characteristic function and after modifying the set of RQFSs. The only remaining step now - is to transform the solved individual \textsc{0-1-Knapsack} problem to OWBQ as it was described in the proof of the statement \ref{st:ldq_np}.

We have to underline that in this case the algorithm \ref{algo:ga} will require some minor changes. Since the number of questions in the LDQ can be less than $n-1$ we will need some method to calculate the intervals' boundaries for the absent subordinal problems. This task can be accomplished e.g. by consequitive splitting of the largest existing interval into two equal ones until the reaching of necessary amount of boundaies. 

%
%
%
%
%
%
\section{Test results}
The results of tests of the algorithms \ref{algo:ls} and \ref{algo:ga} for OWBQ are presented in table \ref{tab:7}, the legend for the header is below: 
\renewcommand{\descriptionlabel}[1]{\hspace{\labelsep}\textsf{#1}}
\begin{description}
\item[Opt.] - Cost of optimal questionnaire
\item[QPF] - Question preference function (see table \ref{tab:4})
\item['Dumb'] - Algorithm \ref{algo:ls} with 'Dumb' RQSFs only
\item[Greed] - Algorithm \ref{algo:ls} with greedy RQSFs from table \ref{tab:4} 
\item[Mixed] - Algorithm \ref{algo:ls} with both 'Dumb'and greedy RQSFs
\item[GA] - Algorithm \ref{algo:ga}
\end{description}

\begin{table}[ht]
\begin{center}
\newcolumntype{d}{D{.}{.}{4}}
\begin{tabular}{|r||d|d|d|d|d|d|} \hline
\multicolumn{1}{|c||}{\textsf{Test No.}} &
\multicolumn{1}{c|}{\textsf{Opt.}} &
\multicolumn{1}{c|}{\textsf{QPF}} &
\multicolumn{1}{c|}{\textsf{'Dumb'}} &
\multicolumn{1}{c|}{\textsf{Greedy}} &
\multicolumn{1}{c|}{\textsf{Mixed}} &
\multicolumn{1}{c|}{\textsf{GA}} \\  \hline\hline
0 & 11.2922 & 11.2922 & 11.4269 & 11.2922 & 11.2922 & 11.2922 \\ \hline
1 & 10.6628 & 10.6628 & 10.9959 & 10.6628 & 10.6628 & 10.6628 \\ \hline
2 & 8.0480 & 8.0526 & 8.7146 & 8.0480 & 8.0480 & 8.0480 \\ \hline
3 & 19.1331 & 19.1373 & 21.5267 & 19.1331 & 19.1331 & 19.1331 \\ \hline
4 & 19.1192 & 19.1758 & 19.9406 & 19.1758 & 19.1758 & 19.1192 \\ \hline
5 & 13.5272 & 13.9316 & 14.5345 & 13.9316 & 13.5272 & 13.5272 \\ \hline
6 & 11.4206 & 11.5124 & 13.6514 & 11.5124 & 11.4966 & 11.4206 \\ \hline
7 & 9.2753 & 9.3732 & 9.3584 & 9.3732 & 9.3337 & 9.2753 \\ \hline
8 & 4.8487 & 4.8487 & 4.9891 & 4.8487 & 4.8487 & 4.8487 \\ \hline
9 & 8.7166 & 8.8975 & 9.9424 & 8.8975 & 8.8975 & 8.8889 \\ \hline
10 & 22.3906 & 22.3906 & 22.8125 & 22.3906 & 22.3906 & 22.3906 \\ \hline
11 & 6.8837 & 6.8837 & 7.3846 & 6.8837 & 6.8837 & 6.8837 \\ \hline
12 & 11.6289 & 11.7187 & 14.3230 & 11.7187 & 11.7187 & 11.7187 \\ \hline
13 & 10.5692 & 10.7699 & 11.4666 & 10.7699 & 10.7699 & 10.7699 \\ \hline
14 & 6.4004 & 6.4711 & 6.7378 & 6.4711 & 6.4711 & 6.4711 \\ \hline
15 & 8.2255 & 8.2322 & 9.2537 & 8.2255 & 8.2255 & 8.2255 \\ \hline
16 & 8.4732 & 8.6399 & 10.7886 & 8.6399 & 8.6399 & 8.5652 \\ \hline
17 & 7.0719 & 7.0838 & 7.5418 & 7.0838 & 7.0838 & 7.1136 \\ \hline
18 & 5.8478 & 5.9778 & 6.5273 & 5.9778 & 5.8478 & 5.8478 \\ \hline
19 & 7.9106 & 8.0136 & 8.8704 & 7.9971 & 7.9971 & 7.9971 \\ \hline
20 & 5.6769 & 5.6769 & 5.6769 & 5.6769 & 5.6769 & 5.6769 \\ \hline
21 & 10.7228 & 10.7228 & 10.9216 & 10.7228 & 10.7228 & 10.7228 \\ \hline
22 & 9.1514 & 9.3193 & 9.3742 & 9.3193 & 9.2781 & 9.1894 \\ \hline
23 & 8.5325 & 8.7274 & 8.5325 & 8.7274 & 8.7274 & 8.6164 \\ \hline
24 & 17.4296 & 17.7087 & 18.5848 & 17.7087 & 17.7087 & 17.7087 \\ \hline
25 & 19.9393 & 20.0682 & 20.3346 & 20.0682 & 20.0682 & 20.0682 \\ \hline
26 & 9.8244 & 9.8244 & 10.8875 & 9.8244 & 9.8244 & 9.8244 \\ \hline
27 & 21.5069 & 21.8195 & 21.8365 & 21.7668 & 21.7668 & 21.6252 \\ \hline
28 & 20.7361 & 20.8671 & 21.4715 & 20.8671 & 20.8671 & 20.8671 \\ \hline
29 & 18.1919 & 18.2026 & 18.6891 & 18.2026 & 18.2026 & 18.2026 \\ \hline
30 & 8.0279 & 8.0279 & 8.3099 & 8.0279 & 8.0279 & 8.0279 \\ \hline
31 & 10.9867 & 11.0133 & 11.1175 & 11.0133 & 10.9867 & 10.9867 \\ \hline
$\sum$ & 3372.1717 & 375.0436 & 396.5232 & 374.9590 & 374.3014 & 373.7154 \\ \hline
\end{tabular}
\end{center}
\caption{Test results\label{tab:7}}
\end{table}

%
%
%
%
%
%
\section{Conclusion}
The proposed approach has let us to develop local search and genetic algorithms which exceed all known approximation algorithms in quality. Because of its universality the developed algorithms can be applied in addition to various  questionnaire optimization problems, also to the \textsc{Minimum Set Cover} and \textsc{Weighted Set Cover} problems, to the \textsc{0-1-Knapsack} problem and probably to other combinatorial optimization problems. LDQ and latticoids are two example of the flexibility of the mathematical model of the questionnaire which let us believe that many other combinatorial optimization problems can be represented as questionnaires and thus can be solved using the proposed approach.

All these problems are characterized by known difficulties in developing a neighborhood function for local search, as well as in the development of genetic operators. The reason for this situation is the specific structure of solutions of all these problems that do not allow efficient implementation of the necessary manipulations. The proposed method gives a relief for this problem.

As local search algorithms and genetic algorithms are very highly adaptive tools and provide the necessary flexibility to be efficient tool in the resolving of different special cases of mentioned above common problems and in the heuristic search for solutions for specific individual problems. 

\bibliographystyle{alpha}
\bibliography{arxiv}

\end{document}